# Data-Driven Breakthroughs and Future Directions in AI Infrastructure: A Comprehensive Review

1st Beyazit Bestami Yuksel
Computer Engineering
Istanbul Technical University
Istanbul, Turkey
yukselbe18@itu.edu.tr

2nd Ayse Yilmazer
Computer Engineering
Istanbul Technical University
Istanbul, Turkey
yilmazerayse@itu.edu.tr

*Abstract*

*This paper presents a comprehensive synthesis of major breakthroughs in artificial intelligence (AI) over the past fifteen years, integrating historical, theoretical, and technological perspectives. It identifies key inflection points in AI's evolution by tracing the convergence of computational resources, data access, and algorithmic innovation. The analysis highlights how researchers enabled GPU-based model training, triggered a data-centric shift with ImageNet, simplified architectures through the Transformer, and expanded modeling capabilities with the GPT series. Rather than treating these advances as isolated milestones, the paper frames them as indicators of deeper paradigm shifts. By applying concepts from statistical learning theory such as sample complexity and data efficiency, the paper explains how researchers translated breakthroughs into scalable solutions and why the field must now embrace data-centric approaches. In response to rising privacy concerns and tightening regulations, the paper evaluates emerging solutions like federated learning, privacy-enhancing technologies (PETs), and the data site paradigm, which reframe data access and security. In cases where real-world data remains inaccessible, the paper also assesses the utility and constraints of mock and synthetic data generation. By aligning technical insights with evolving data infrastructure, this study offers strategic guidance for future AI research and policy development.*

## I. INTRODUCTION

Artificial intelligence (AI) technologies have advanced rapidly and dramatically over the past decade, reshaping not only technical methodologies but also the foundational paradigms of data processing. Early developments in AI gained momentum primarily through improvements in computational power. Over time, however, researchers increasingly relied on larger datasets and more efficient algorithms to drive progress. Within this evolutionary trajectory, distinct breakthroughs redirected the focus of AI research and revealed which technologies and approaches could remain sustainable.

Forecasting the future trajectory of AI extends beyond academic interest. It serves as a key driver for strategic direction in industry, policy development, and career trajectories across multiple fields. As the costs of acquiring data and maintaining compute infrastructure continue to rise, researchers and decision-makers must identify the forces behind the next transformative leap. Revisiting historical breakthroughs does more than offer retrospective understanding it enables researchers to build grounded foresight by recognizing recurring patterns of innovation.

This study traces the history of AI breakthroughs and investigates the structural shifts that enabled them. The analysis follows three primary dimensions: computational resources (compute), the quantity and nature of data (data), and algorithmic innovation. By applying principles from statistical learning theory particularly concepts like sample complexity and number of samples, the study explains why certain advances qualified as genuine breakthroughs.

The paper also examines current challenges in the AI landscape. As access to open data sources declines and privacy regulations tighten control over private datasets, researchers face increasing constraints. In response, the study evaluates emerging paradigms such as federated learning, privacy-enhancing technologies (PETs), and data site infrastructures. It outlines a unified framework to understand how ethical, technical, and policy considerations will shape the future of AI infrastructure.

## II. STATISTICAL LEARNING THEORY AND THEORETICAL FOUNDATIONS

One of the most fundamental ways to understand how and why artificial intelligence (AI) systems learn involves evaluating this process through the lens of Statistical Learning Theory (SLT). This framework offers a conceptual and mathematical foundation for estimating the accuracy a model can achieve based on the number of samples it receives. Two primary factors govern the quality of learning: the number of samples and the sample complexity.

SLT starts with an intuitive premise: increasing the amount of data enhances a model's capacity to learn. For instance, the MNIST dataset contains 60,000-labeled images of handwritten digits [1]. If we could construct a dataset that covers all possible 28×28 pixel permutations amounting to a space of $2^{784}$ images the model would effectively "see" every possible digit, thereby approaching a near-perfect classifier. However, real-world conditions render this infeasible. Data remains finite, so the system must infer patterns from examples it has never encountered. This brings sample complexity into focus—the measure of data required by an algorithm to reach a targeted performance threshold [2]. Algorithms with low sample complexity achieve high performance with relatively few "examples, whereas those with higher complexity require significantly larger datasets to perform comparably. Many studies in contemporary machine learning literature (e.g., NeurIPS and ICML papers) directly or indirectly aim to reduce sample complexity.

In contrast, data efficiency refers to how well an algorithm learns from a limited number of samples. Techniques such as dropout regularization or architectural innovations like attention mechanisms improve data efficiency by enabling richer inference from the same input. The Transformer architecture illustrates this principle concretely. Whereas earlier models like RNNs and CNNs required high sample complexity to capture sequential dependencies, the Transformer uses attention alone to learn long-range relationships with fewer samples.

SLT also emphasizes the importance of data quality and diversity alongside data quantity. Large and diverse datasets generally enhance generalization capability. However, if the algorithm lacks efficiency in terms of sample complexity, these datasets may prove insufficient or prohibitively costly for achieving desired outcomes. A commonly cited heuristic in the field referred to as "napkin math" summarizes this relationship as follows:

AI Capability $\approx$ Number of Samples $\times$ Data Efficiency

This formulation conveys a fundamental insight: more data typically strengthens models, but better algorithms can reach high performance with less. Compute power also plays a decisive role in this equation. With high-capacity hardware, researchers can train on larger datasets. GPUs, in particular, support both the generation of new examples (e.g., via video game simulations) and the processing of those examples. This capability creates larger learning opportunities.

SLT further underscores two key dimensions of data: quantity and quality. Possessing a sufficiently large and diverse dataset increases a model's capacity to generalize. However this concept alone does not guarantee performance. If the underlying algorithm exhibits high sample complexity, even vast quantities of data will not suffice or will impose substantial costs.

Figure 1 visualizes the relationship between sample count and model accuracy. It compares algorithms with low, medium, and high sample complexity. The curves illustrate how algorithms with lower complexity reach higher accuracy with fewer examples, while those with high complexity show slower performance gains.

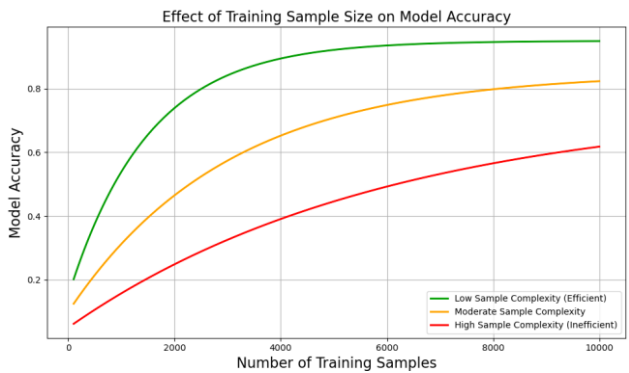

*Figure 1: Relationship between Number of Samples and Model Accuracy*

Statistical Learning Theory offers more than just a conceptual lens—it provides researchers with a practical framework to optimize AI development. It helps them choose effective architectures, define performance limits, and align data requirements accordingly. This perspective clarifies the significance of past breakthroughs and identifies the parameters likely to drive the next wave of progress. By grounding historical milestones in theoretical structure, SLT empowers researchers to interpret progress through a coherent analytical framework. In the next section, we explore these breakthroughs from this viewpoint.

III. HISTORICAL MILESTONES IN AI BREAKTHROUGHS

Some developments in the history of artificial intelligence (AI) represent more than just technical achievements they signify paradigm shifts. These breakthroughs either enabled learning at previously unattainable scales or introduced new approaches that displaced incumbent methods. As illustrated in Figure 2, each of these milestones highlighted below represents a notable leap in AI capabilities. We categorize them across three fundamental dimensions: compute capacity, data scale, and algorithmic efficiency. This framing allows for a more nuanced understanding of why these developments served as turning points.

*2009 – Deep Learning with GPUs (Raina & Ng)*

In 2009, Andrew Ng and his team introduced the use of GPUs for training neural networks [3]. Until then, most models had relied on CPU-based training, which imposed significant time and resource constraints, especially for large-scale models. Their work demonstrated that it was feasible to train networks with over 100 million parameters within reasonable timeframes. This marked a major leap in AI fueled by expanded computational resources.

*2010 – Release of the ImageNet Dataset*

Stanford University introduced the ImageNet dataset in 2010 [4], which, at the time, stood as one of the largest and most comprehensive labeled image datasets in the field. Containing millions of annotated images, ImageNet made comparative benchmarking in visual recognition possible and actively encouraged algorithmic progress. This development underscored the importance of data scale in driving AI advancement.

*2012 – AlexNet and Dropout*

That same year, AlexNet's success in the ImageNet competition redefined what deep learning could achieve at scale. By training the model on GPUs and incorporating Dropout [5] as a regularization technique, researchers demonstrated a practical method to curb overfitting while improving generalization. Dropout, by randomly deactivating units during training, encouraged the model to rely on distributed representations, which in turn reduced its dependence on specific pathways and lowered the sample complexity. This development not only improved performance without increasing data volume but also illustrated how architectural simplicity and data efficiency could reinforce each other. As a result, the community began to recognize that meaningful performance gains could emerge not just from larger models or datasets, but from smarter training techniques that made better use of what was already available.

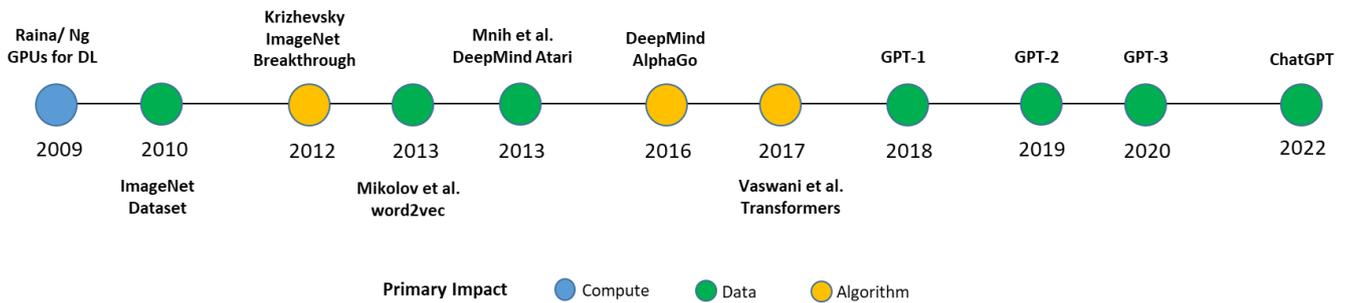

*Figure 2: AI Breakthrough Timeline*

## 2013 – Word2Vec: Uncovering the Power of Simplicity

In 2013, Thomas Mikolov and colleagues introduced Word2Vec [6], a breakthrough in natural language processing. The model became known for its capacity to represent semantic relationships mathematically for instance below.

King - Man + Woman = Queen

However, the true breakthrough lay not just in semantic modeling but also in the ability to train on massive datasets using an extremely simplified architecture.

At the time, dominant approaches based on recurrent neural networks (RNNs) required high sample complexity and came with steep computational costs. Mikolov purposefully stripped down the architecture: he removed non-linear hidden layers, approximated the output layer, and implemented the entire model in optimized C code. This simplification yielded remarkable scalability, enabling the model to train on trillions of words in mere minutes. Andrew Trask, during his participation at ICML 2015, recalled witnessing Mikolov express the following motivation:

*"My main goal was to simplify the algorithm so we could train on much more data than with previous RNN-based approaches."*

From a statistical learning theory perspective, Word2Vec exemplifies a model with low sample complexity that achieved high accuracy by scaling with data volume. The parallel threading method used in its original implementation even violated conventional software engineering norms by allowing simultaneous writes to shared memory yet the model's structure tolerated such disorder, enabling fast and effective training.

Although Word2Vec may appear to be a standard case of algorithmic innovation, its actual contribution lies in the fusion of a deliberately "weak" model with vast quantities of data. In essence:

Weak Model × Large Data = High Performance

This episode reinforces a recurring pattern in AI: access to massive and high-quality data often yields more substantial gains than sophisticated algorithms alone.

## 2015 – DeepMind and Atari Games

The combination of Q-learning and convolutional neural networks in DeepMind's Atari project demonstrated that reinforcement learning (RL) could facilitate the generation of effectively infinite data [7]. The model learned from visual inputs derived from gameplay, simulating experiences that would be difficult to replicate in the physical world. This work established a new direction in synthetic data generation and opened the door to the idea of learning from infinite virtual environments.

## 2016 – AlphaGo: Algorithmic Intelligence Meets Infinite Data Generation

AlphaGo introduced a new paradigm not only in gameplay strategy but also in data generation, learning capacity, and decision-making architecture [8]. Previous systems learned from a finite number of games played by human experts. In contrast, AlphaGo redefined this trajectory. After pretraining on expert games, the system began playing millions of games against itself using a self-play strategy, thereby generating a virtually limitless training corpus. This artificial data abundance enabled a depth of learning that real-world data alone could not provide.

The system's success relied not only on data scale but also on a multilayered algorithmic architecture. AlphaGo's decision-making pipeline integrated three core components:

1) *Monte Carlo Tree Search (MCTS):* This algorithm explored potential moves as a branching tree, computing probabilities and value estimates for each node. While it did not learn directly, it implemented a highly efficient strategy to search for optimal actions.

2) *Convolutional Neural Networks (CNNs):* These networks interpreted visual board states to support move prediction via:

- Policy Network: "Which move should I make in this position?"
- Value Network: "What is the win probability from this board configuration?"

3) *Self-Generated Data:* Each self-played game produced new examples and experiences, continuously enriching the training signal and enabling the model to improve generalization and accuracy over time.

What set AlphaGo apart was not merely its algorithmic design but its ability to pair these methods with a scalable simulation environment that generated training data internally. Unlike traditional machine learning workflows that depend on externally collected data, AlphaGo's environment served as both model and data engine.

AlphaGo marked a decisive moment in the history of AI for two reasons: it pushed the boundaries of data generation and combined algorithmic depth with modular, interconnected components. The breakthrough did not lie in compute or architecture alone but in the cohesive alignment of data, algorithms, and hardware. AlphaGo remains a rare instance where both "more data" and "better algorithms" contributed synergistically to a transformative result.

*2017 – Transformer*

The Transformer architecture, introduced in "Attention is All You Need," replaced complex RNN and CNN models with a streamlined yet powerful alternative [9]. The architecture relied exclusively on attention mechanisms to capture long-range dependencies, eliminating recurrence. This approach significantly reduced sample complexity and enabled efficient training on massive datasets. Researchers widely recognized Transformer not only for its architectural simplicity but also for its ability to achieve high data efficiency at scale making it a foundational innovation in modern AI.

*2018–2020 – The GPT Series: Generalization Through Scale and Data-Driven Breakthroughs*

OpenAI's Generative Pre-trained Transformer (GPT) series stands among the most compelling achievements in deep learning. The core success of these models stemmed from their ability to scale the Transformer architecture to handle immense data volumes. GPT models adopted a two-phase training strategy: first, large-scale self-supervised pretraining on unlabeled text to develop general-purpose linguistic representations; then, fine-tuning on task-specific datasets using smaller, supervised corpora.

GPT-1 introduced this paradigm with a straightforward proposition [10]:

"Labeled data is scarce, but unlabeled text is abundant. We first perform large-scale self-supervised pretraining and then apply task-specific fine-tuning on small datasets."

This strategy enabled the model to leverage massive data sources effectively and achieve strong performance across a broad set of tasks. GPT-1 did not introduce a novel architecture but rather demonstrated how data scale could elevate an existing one.

GPT-2 followed the same principle [11]. It trained on WebText a massive dataset drawn from millions of web pages. This was not an architectural leap but a demonstration of how increasing data volume could directly enhance model performance. More data translated into better generalization and higher accuracy.

GPT-3 emphasized scale even further. Although the authors did not explicitly frame the achievement as a "data breakthrough," [12] the model's size and parameter count implicitly revealed its relationship with data. GPT-3 outperformed earlier models not merely due to parameter growth but because it could meaningfully utilize more and more diverse data. This progression showed that scaling up a model also required a proportional increase in training data to unlock additional learning capacity.

At this point, a critical balance emerged. Enlarging model capacity without increasing data volume risked overfitting. Success depended on scaling both the model and the data in tandem. The GPT series demonstrated that harmony between architecture and data quantity was essential not optional for effective learning at scale.

GPT's overall success did not arise from architectural novelty alone but from aligning data scale with model capacity. This alignment enabled the models to perform zero-shot and few-shot learning across a wide range of tasks. The series provided definitive evidence that "more data enables better generalization" and showed how to translate that principle into state-of-the-art performance in real-world AI applications.

*2022 – ChatGPT*

Built on GPT-3.5, ChatGPT marked a pivotal moment in both technical infrastructure and user experience. OpenAI integrated Reinforcement Learning from Human Feedback (RLHF) to enable the model to generate more natural, context-aware responses [13]. By deploying ChatGPT through a publicly accessible conversational interface, OpenAI expanded the societal reach of AI and catalyzed its cultural integration. This model embodied a balanced synthesis of compute, data, and algorithmic design, positioning it as a multidimensional breakthrough.

Viewed collectively, these breakthroughs reflect a consistent pattern where algorithmic design, data availability, and compute capacity evolved in tandem. Researchers leveraged GPU acceleration not only to expand model depth and parameter counts, but also to accelerate data throughput, enabling iterative experimentation at an unprecedented scale. As compute infrastructure matured, it facilitated the emergence of training paradigms that aligned more closely with real-world complexity supporting larger datasets, more nuanced architectures, and longer training cycles. In this ecosystem, computational power shaped the way researchers formulated problems, designed experiments, and optimized model performance across diverse domains.

Similarly, the ImageNet milestone did not hinge on dataset scale alone it also incorporated algorithmic interventions such as Dropout. This simple yet powerful technique mitigated overfitting by injecting stochastic noise into input and intermediate layers, thereby encouraging generalization. In effect, Dropout produced a data augmentation effect by training the model to behave as if it had encountered a wider diversity of examples. In other words, Dropout amplified existing data representations through implicit diversification.

The ascent of deep learning owes much to its architectural simplicity. While probabilistic graphical models (PGMs) provide rich theoretical frameworks and expressive capabilities, they demand significant computational resources that limit practical scalability. In contrast, deep learning architectures align more naturally with GPU acceleration, enabling faster execution and better adaptation to high-volume data environments. This alignment allowed deep learning models to take full advantage of the computational and data resources available during the GPU-driven phase of AI development. Despite their conceptual depth, PGMs found limited success in delivering scalable performance under real-world data conditions, whereas deep learning systems demonstrated consistent gains through streamlined design and efficient computation.

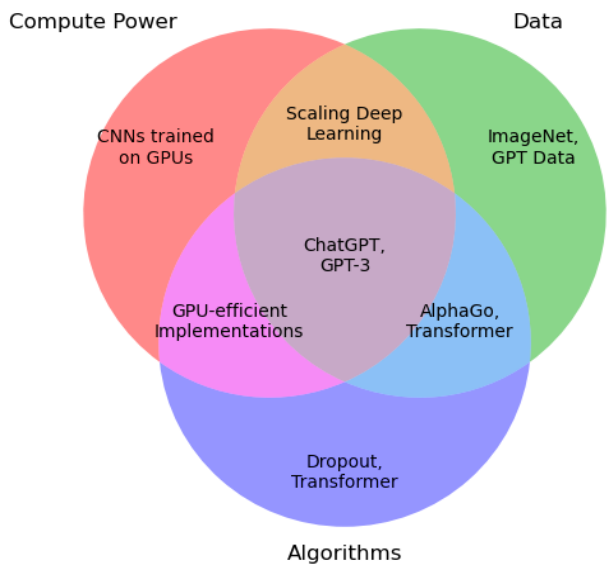

*Figure 3: Factors and Overlaps in AI Breakthroughs*

A closer look at major breakthroughs in AI reveals that most innovations arose from the interplay of compute power, algorithmic design, and data availability. However, historical analysis suggests that access to high-quality data consistently played a dominant role. The Venn diagram in Figure 3 illustrates the conceptual intersections where significant AI breakthroughs occurred. For example, models like GPT-3 and ChatGPT did not evolve solely due to advanced architectures or hardware they relied fundamentally on access to vast corpora of textual data. Similarly, the ImageNet era demonstrated how large-scale, labeled datasets could trigger paradigm shifts.

This historical pattern suggests a critical insight: data access has acted as the primary catalyst, while hardware and algorithms evolved to accommodate the growing demands of data-intensive learning. We conclude that:

*"Engineered intelligence may yield short-term gains, but long-term advancements arise from scalable learning algorithms applied to increasingly powerful compute and increasingly rich datasets."*

Therefore, this study supports the projection that the next significant breakthrough in AI will likely stem from improved access to larger, higher-quality, and more inclusive data resources.

## IV. WHERE WILL THE NEXT BREAKTHROUGH COME FROM?

When we examine the major AI breakthroughs of the past fifteen years, we find that two core drivers underpinned most of them: the scaling of data and advances in computational power. While researchers occasionally introduced structural innovations in algorithms, most improvements focused on utilizing larger datasets more effectively. For instance, the progress observed in the GPT series resulted not from radical architectural change but from scalable data usage and expanded pretraining strategies.

As Andrew Trask, a leading AI researcher, emphasized [18], anyone aiming to anticipate the next wave of breakthroughs must begin with this question:

"Over the next 5–10 years, how can we increase the available data by a factor of 10 to 1000?"

This question transcends technical curiosity. It outlines a strategic vision that will shape the future direction of AI research.

**Talent, Hardware, or Data?**

Training researchers capable of building advanced algorithms takes years. Pioneers like Geoffrey Hinton, Yann LeCun, and Yoshua Bengio shaped the field through decades of cumulative contributions. However human talent grows at a natural limit even a 10% annual increase would represent an optimistic estimate. As such, breakthroughs rooted in algorithmic novelty will likely progress too slowly to drive near-term leaps [14].

Similarly, compute performance faces physical and economic ceilings. Moore's Law is approaching saturation, and GPU throughput typically increases by just 2-4x per year. Although companies like NVIDIA offer scalable cloud-based infrastructure, a 1000x increase in computational throughput remains unlikely in the short term. Promising technologies like quantum computing and analog computation have not yet reached operational or commercial maturity [15].

**The Key to the Next Breakthrough: Data Volume and Access**

Given these constraints, future breakthroughs will most likely originate not from better algorithms or hardware, but from data itself. However, success will not depend solely on more data it will require new types of data, novel access protocols, and the ability to analyze previously inaccessible or sensitive datasets.

Approaches such as federated learning, privacy-preserving computation, and synthetic data generation offer promising pathways to unlock these new data regimes [16]. These techniques help researchers expand the usable data ecosystem while preserving privacy and ensuring scalability.

In this light, we can state the following with confidence:
The next significant breakthrough in AI will stem from leveraging larger, more diverse, and more accessible datasets through well-designed utilization strategies, rather than relying solely on advances in algorithmic power.

**A Policy and Strategy Perspective**

This insight extends beyond technical forecasting. It offers concrete strategic guidance for both public and private stakeholders. If governments and institutions implement data-centric public policies such as open data standards, interoperable sharing protocols, and privacy-preserving infrastructure then researchers will gain the tools to develop simpler yet powerful models that reduce sample complexity.

Therefore, the central thesis of this article "More data leads to stronger AI" does not merely describe the past. It articulates a core principle that will define the next frontier of AI development.

## V. TRANSITIONING TOWARD DATA-CENTRIC APPROACHES IN AI

Artificial intelligence continues to evolve through a fundamental shift in its guiding paradigm. While earlier advancements gained momentum through expanded computational capacity and algorithmic improvements, current progress increasingly depends on data as the central driving force. The effectiveness of a system now hinges on how it harnesses computational resources and, more importantly, how it integrates data into the learning process. This transition toward data-centric AI marks a pivotal transformation one that defines the direction of future development rather than offering a mere alternative.

Open data sources have begun to diminish significantly. Many platforms such as Reddit, X (formerly Twitter), and news websites have started restricting access to their content or closing it off entirely. These restrictions directly limit the diversity and volume of large-scale training datasets harvested from the web. Although synthetic data generated through simulation environments (such as visual data from video games) can prove effective in some domains, they fall short in fields that require real-world complexity and contextual nuance such as natural language processing, biomedical analysis, and socioeconomic modeling.

In this context, private datasets such as hospital records, internal corporate documents, financial transactions, and government archives represent the most valuable and meaningful data sources for future AI breakthroughs. Yet accessing and utilizing these datasets introduces not only technical but also ethical, legal, and political challenges. Regulatory frameworks like the European Union's General Data Protection Regulation (GDPR), Turkey's KVKK, and similar data protection laws have placed considerable constraints on data sharing. In parallel, institutions face heightened concerns about competitive risk, data leakage, and public trust, which further reinforce restrictive data policies.

At this crossroads, the data-centric mindset redefines AI not as a quest to accumulate more data at any cost, but as an effort to access data ethically, securely, and under regulatory oversight. We now stand at the threshold of a shift from the age of open data to the age of private data. The new imperative is to center data in AI workflows without centralizing it physically. This requirement compels the development of new technological infrastructures and responsible access mechanisms.

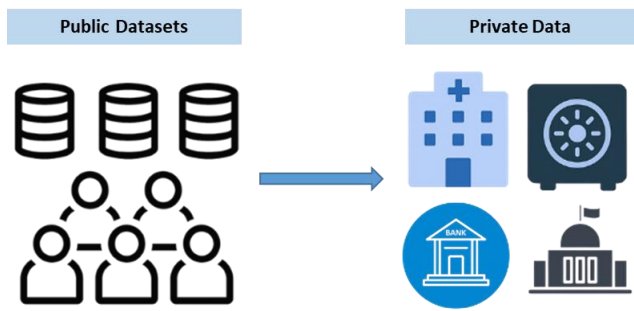

*Figure 4: Data Ecosystem Evolution Diagram*

Figure 4 illustrates this transformation. On the left side of the diagram, open data repositories (e.g., publicly available datasets) represent the past; on the right, private data vaults (e.g., hospital systems, enterprise platforms) symbolize the future. The arrow connecting these domains conveys the transition one that emphasizes secure, controlled, and privacy-preserving access.

This transition demonstrates that AI must focus not only on the quantity of data but also on its quality and contextual meaning. Open data often lacks structure, completeness, or contextual richness. In contrast, private data tends to be more consistent, labeled, and semantically dense factors that significantly influence model performance. Table 1 presents a comparative view of open versus private data sources:

*Table 1: Comparative view of Open versus Private data sources*

| Attribute | Open Data | Private Data |
|---|---|---|
| **Accessibility** | Decreasing; subject to platform restrictions | Generally abundant within specific domains |
| **Data Quality** | Varied; often noisy and unstructured | Higher accuracy; context-rich and labeled |
| **Legal/Ethical Risks** | Low; minimal compliance requirements | High; demands strict adherence to privacy regulations |
| **Access Control** | Open or platform-limited | Strict; controlled by data owners |

As a result, the data-centric AI paradigm calls for an interdisciplinary framework. Engineering knowledge alone no longer suffices. Fields such as law, ethics, public policy, and the life sciences must become active collaborators in shaping AI systems of the future.

## VI. EMERGING SOLUTIONS FOR DATA PRIVACY AND ACCESS IN AI

Data serves as the foundation of AI systems, yet its practical value depends on the ability to use it effectively. In fields that involve sensitive information—such as healthcare, finance, and enterprise operations—data usage brings substantial challenges related to privacy, legal compliance, and cybersecurity. As AI development progresses, its success increasingly depends on managing not just data access, but also the context, location, and governance under which that access occurs. In response, researchers have proposed a range of solutions designed to address these challenges at their core.

### *Federated Learning*

Federated learning enables model training without transferring raw data to a central server. Instead, each data holder (e.g., a hospital, financial institution, or individual device) updates the model locally using its own data. It then shares only the updated model parameters. A central coordinator aggregates these local updates to produce a global model.

The process that demonstrated it Figure 5, ensures that sensitive data never leaves its origin. It preserves user privacy and facilitates compliance with legal frameworks by design.

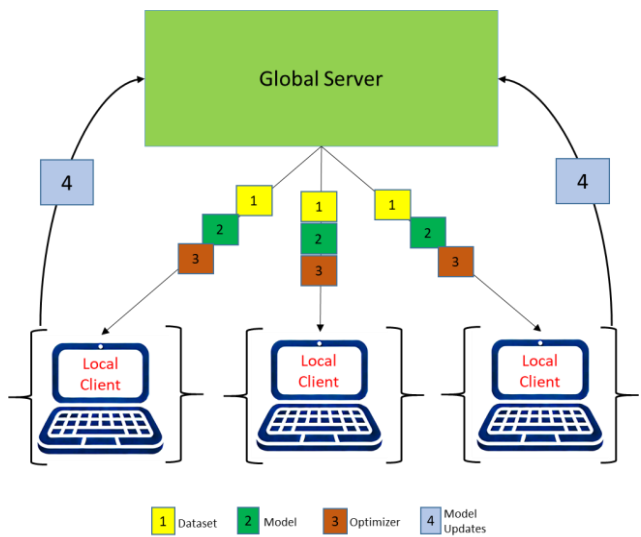

*Figure 5: Federated Learning Model Training Process*

This approach proves especially effective in distributed yet privacy-sensitive environments such as mobile devices, IoT systems, and hospital networks. However, the success of federated learning depends on several factors: the heterogeneity of data distributions, the quality of local model updates, and the integrity of secure aggregation methods. As such, federated learning demands a new mindset one that rethinks conventional algorithmic design.

*Privacy-Enhancing Technologies (PETs)*

PETs encompass a suite of technologies designed to guarantee privacy throughout the data processing pipeline. This category includes advanced methods such as differential privacy, homomorphic encryption, and secure multi-party computation (SMPC).

Homomorphic encryption allows computations to run directly on encrypted data. Users can extract insights without ever exposing the raw content. SMPC enables multiple parties to perform joint computations without revealing their individual datasets. These methods are no longer purely theoretical constructs. In domains that demand stringent compliance such as healthcare, defense, and finance researchers have begun deploying PETs as viable, real-world solutions.

*DataSite Paradigm*

The DataSite Paradigm, developed by Andrew Trask and the OpenMined team [19], reimagines how researchers access data. Inspired by classical web architecture but redesigned with data privacy as its core, this paradigm prevents direct downloads of raw datasets. Instead, researchers send their code to the data site. After undergoing inspection and approval by the data owner, the code executes locally on the server. Only the computed output not the raw data returns to the researcher.

This method adheres to the principle: "Send the code to the data, not the data to the code." Rather than transferring sensitive datasets, researchers send models or queries to where the data resides. This structure preserves data locality while enabling secure and accountable analysis.

Frameworks like PySyft [20] support this workflow, integrating PETs with functionalities such as remote execution, audit logging, project governance, and automated policy enforcement. For instance, when a researcher submits code to a data site, the system can automatically verify compliance with predefined data policies removing the burden of manual review from the data steward. This automation enhances both security and scalability.

*Network Logic and the DNS Analogy*

Data sites do not function in isolation they belong to a broader data network. We compare this structure to the Domain Name System (DNS) just as DNS resolves domain names into machine-readable addresses, a data site network enables researchers to discover, access, and interact with a decentralized array of private datasets.

Figure 6 outlines the core operational logic:

- The data owner uploads the dataset to a local server and defines access protocols.
- The researcher submits analytical code to the data site.
- The code undergoes manual or automated review.
- Upon approval, the system executes the code on the local server.
- The researcher receives either the summary output or a distilled version of the model.

*Algorithm 1: Data Site Code Flow*

```
START
// 1. Data owner uploads the data to the system
DataOwner.upload(data)
DataOwner.defineAccessProtocol(protocol)
// 2. Researcher submits their code
Researcher.submitCode(code)
// 3. Code is reviewed
IF isCodeCompliant(code) == TRUE THEN
    // 4. Code is executed on the local server
    result = executeLocally(code, data)
    // 5. Summary result is sent to the researcher
    sendToResearcher(resultSummary(result))
ELSE
    REJECT code WITH explanation
ENDIF
END
```

Throughout this process in Algorithm 1, the system logs every step, and no analysis proceeds without the explicit approval of the data owner. This structure ensures a transparent, auditable, and scalable framework.

**Use Case: DataSite in Healthcare**

Consider a hospital that deploys a DataSite on its own infrastructure and uploads patient health records into the system. The Ministry of Health sends the following query to the DataSite:

*result = df[(df["age"] > 65) & (df["diagnosis"] == "heart attack")].shape[0] / df.shape[0]*

This code calculates the proportion of elderly patients diagnosed with heart attacks. Once the data owner reviews and approves the code, the system executes it locally. The researcher receives only the computed result such as "12.4%" without accessing any raw patient data. This approach ensures data remains on-site, analysis proceeds securely, and both ethical and legal compliance are maintained.

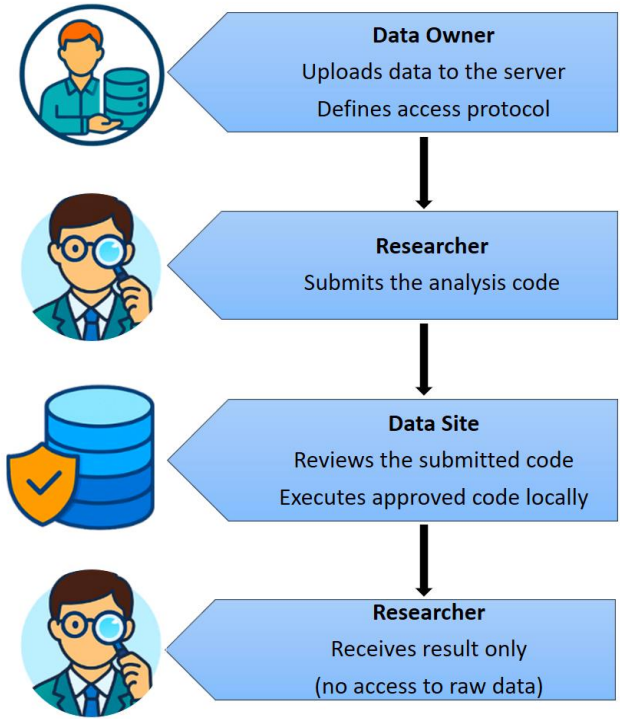

*Figure 6: Data Site Paradigm Flow Diagram*

This approach shown in Table 2, provides a well-structured foundation for compliance reporting and supervisory technologies in highly regulated sectors such as public administration, finance, and healthcare. It enables organizations to conduct audits and data analyses without ever extracting raw data. Institutions can thus preserve their data sovereignty while contributing to scientific progress.

*Table 2: Comparison of Traditional and Modern Approaches*

| Attribute | Traditional Centralized Model | Federated Learning / DataSite Approach |
|---|---|---|
| Data Movement | Transfers data to a centralized server | Keeps data in place; transfers code or model weights |
| Privacy Risk | High (data may be exposed) | Low (raw data remains local) |
| Regulatory Compliance | Often problematic | Aligned with GDPR, KVKK, and similar frameworks |
| Control Distribution | Centralized authority | Distributed; operates with data owner's explicit consent |

In conclusion, solutions that balance data privacy and data access do not merely represent technical optimizations they constitute foundational shifts that will shape the future of AI. These frameworks not only ensure legal compliance but also support scalable, trustworthy collaboration across institutions and even national borders.

### VII. Mock and Synthetic Data: Practical Applications and Limitations

Up to this point, we have argued that building artificial intelligence systems involves much more than designing algorithms or training models. The type, quality, and accessibility of data used in this process play a decisive role in shaping a model's performance. However, in domains where direct access to real data proves infeasible such as healthcare, finance, or user-specific systems researchers and developers frequently turn to two practical alternatives: mock data and synthetic data.

**Mock Data:** Mock data refers to artificial datasets that replicate the structural characteristics of real data while containing randomly generated or anonymized values. Developers typically use mock data to test system architectures, verify API endpoints, validate visualization pipelines, or ensure user interfaces run without crashing.

Common use cases for mock data include:
- Preventing system failures during early-stage software testing,
- Prototyping applications before real data becomes available,
- Observing interface behavior when data access is restricted.

However, mock data generally lacks semantic content and fails to reflect real-world dynamics. For instance, mock ECG signals do not encode medically relevant patterns such as heart rhythm, waveform morphology, or frequency. As a result, mock data has limited application and primarily supports early development phases rather than model training or evaluation.

**Synthetic Data:** Synthetic data is artificially generated by algorithms that model the statistical properties of a specific data type. Although it mimics real-world structure, synthetic data does not correspond to any actual user or event. When it maintains statistical similarity to real data, synthetic data can effectively support training, validation, and performance benchmarking.

Key advantages of synthetic data include:
- The ability to approximate real-world outcomes,
- Reduced privacy risk,
- Generation of diverse datasets for various use cases,
- Shareability and publication under ethical frameworks.

For example, if a hospital cannot release actual patient records, researchers may create a synthetic dataset that matches the statistical distributions of the hospital's data. This allows ongoing development and testing of models without violating privacy. In domains like ECG signal processing, MRI analysis, or biometric sensor data, mathematically modeled synthetic signals make it possible to develop high-performing models while maintaining ethical safeguards.

Nevertheless, synthetic data has several limitations:
- It may fail to capture the full range of variance found in real-world events,

- Approaching hyper-realism can introduce re-identification risk,
- Its generation process can be computationally intensive and complex.

*Table 3: Comparison of Mock Data and Synthetic Data*

| Attribute | Mock Data | Synthetic Data |
|---|---|---|
| Data Source | Randomly generated, template-based | Statistical properties derived from real data |
| Purpose | Testing, prototyping, system validation | Model training, analysis, scenario simulation |
| Realism | Low | Medium to high |
| Privacy Risk | Very low | Moderate (if overly similar to real data) |

The key distinction between mock and synthetic data (Table 3) lies in their purpose and stage of application. Developers use mock data during early development phases to bypass system-level issues, while synthetic data supports model learning even in the absence of real datasets providing higher functional value.

Today, with the increasing adoption of PETs and federated learning, synthetic data generation has become a strategic method at the enterprise level. For instance, tools such as ECGSYN, developed at MIT [17], generate synthetic ECG signals that mimic real cardiac activity. These tools enable medical research to proceed without encountering legal or ethical roadblocks.

## VIII. IMPLICATIONS FOR FUTURE AI INFRASTRUCTURE AND RESEARCH DIRECTIONS

The evolution of artificial intelligence affects far more than just system architectures. It also influences how developers build, deploy, and govern AI systems. Future breakthroughs will require more than just powerful algorithms or large datasets. They must emerge from infrastructures that are secure, scalable, and politically sustainable.

*The Source of Future Breakthroughs: Data Quality and Security*

Training large-scale models once posed the central challenge in AI. Today, the critical bottleneck lies in accessing high quality, ethically usable data. With diminishing availability of open data and increasing restrictions around institutional datasets, researchers now face direct limitations. As a result, future breakthroughs will depend not only on how much data systems can use, but also on how that data is accessed and processed.

Technologies such as PETs, federated learning, and DataSite architectures offer foundational responses to this challenge. Yet their success hinges not only on technical feasibility but also on regulatory approval, institutional readiness, and public trust. In this regard, public policy, civil society initiatives, and the open-source movement will all play defining roles.

*Transformation in Research and Industrial Practice*

These emerging infrastructures redefine the research pipeline itself. Instead of downloading datasets and training models on centralized servers, researchers must now adopt distributed, code-based access methods. This shift requires a full rethinking of algorithm design, evaluation procedures, and deployment strategies.

In industry, the same transformation will drive the emergence of data-local, privacy-preserving business models. Especially in sectors like healthcare, finance, and law where data sharing becomes nearly impossible companies will need to run AI workloads within the data owner's environment.

This trend will inevitably lead to new institutional investments. Future AI ecosystems will depend on advanced GPU clusters, private data vaults, federated-compatible device networks, and PET-enabled software stacks. In the long term, quantum computing and analog computation may further scale these infrastructures and push the boundaries of what AI systems can achieve.

*Future Research Directions*

Future research in AI will likely concentrate on three key domains:

- Improving Federated Learning Algorithms: Researchers will aim to enhance efficiency and stability for systems operating across heterogeneous devices, non-iid data distributions, and limited compute environments.
- Advancing PET Frameworks: Work will focus on developing lighter, faster, and more deployable variants of techniques such as homomorphic encryption, differential privacy, and SMPC. This effort includes both algorithmic optimization and software-hardware integration.
- Enhancing Realism in Synthetic Data Generation: Researchers will explore domain-specific generative models that mimic real data without posing re-identification risks. Techniques such as GANs, VAEs, and physics-informed simulators will lead this evolution.

The next generation of AI infrastructure must prioritize data security, uphold ethical principles, operate in technically distributed environments, comply with legal frameworks, and remain socially transparent. Achieving this vision will require deep coordination across academia, industry, and regulatory bodies ensuring that breakthroughs remain technically impactful and socially sustainable.

## IX. CONCLUSION

This article examined the evolution of artificial intelligence over the past fifteen years from a multidimensional perspective, systematically evaluating the core driving forces: computational capacity, data volume, and algorithmic innovation. Each breakthrough, when viewed through a historical lens, represented not just a technical advancement but also a structural paradigm shift that redefined the trajectory of AI research.

The journey that began in 2009 with the introduction of GPUs for deep learning, continued through the data abundance enabled by ImageNet, the deep learning revolution sparked by AlexNet, and the peak of algorithmic efficiency brought by the Transformer architecture. Models like the GPT series and ChatGPT demonstrated how self-supervised learning on massive datasets could lead to transformative results.

These developments highlight a key pattern: AI breakthroughs frequently arose where abundant data aligned with advances in learning algorithm efficiency. However, with open data resources dwindling and private data becoming increasingly inaccessible, sustaining future breakthroughs will require a shift in data infrastructure strategies.

Strategic priorities for the future should focus on the following:

- Building ethical and policy frameworks that define who uses data, how, and under what conditions ensuring transparency, accountability, and inclusiveness,
- Expanding collaboration protocols between researchers and institutions to facilitate secure data access and sharing,
- Evaluating algorithms not only on accuracy but also on fairness, security, and sustainability.

In conclusion, the future of AI does not rest solely on greater computational power or larger models. It depends on better-organized, secure, ethical, and accessible data ecosystems. Establishing these ecosystems will require multidisciplinary collaboration, bringing together engineering, law, ethics, the social sciences, and public policy as integral components. The most critical factor in steering the next wave of AI breakthroughs will no longer be just "how much data we have," but "how we use data ethically, securely, and in the right context."